\documentclass{article}

\usepackage{PRIMEarxiv}

\usepackage[utf8]{inputenc} 
\usepackage[T1]{fontenc}    
\usepackage{hyperref}       
\usepackage{url}            
\usepackage{booktabs}       
\usepackage{amsfonts}       
\usepackage{nicefrac}       
\usepackage{microtype}      
\usepackage{lipsum}
\usepackage{fancyhdr}       
\usepackage{graphicx}       

\usepackage{times}
\usepackage{soul}
\usepackage[small]{caption}
\usepackage{amssymb}
\usepackage{amsmath}
\usepackage{amsthm}
\usepackage{algorithm}
\usepackage{algorithmic}
\usepackage{wrapfig}
\usepackage{breqn}

\graphicspath{{media/}}     

\title{InferNet for Delayed Reinforcement Tasks: \\Addressing the Temporal Credit Assignment Problem
}
\author{
Markel Sanz Ausin, Hamoon Azizsoltani, Song Ju, Yeo Jin Kim, Min Chi \\
North Carolina State University \\
\{msanzau, mchi\}@ncsu.edu \\
}

\begin{document}
\maketitle

\begin{abstract}
The temporal Credit Assignment Problem (CAP) is a well-known and challenging task in AI. While Reinforcement Learning (RL), especially Deep RL, works well when immediate rewards are available, it can fail when only delayed rewards are available or when the reward function is noisy. In this work, we propose delegating the CAP to a Neural Network-based algorithm named \emph{InferNet} that explicitly learns to infer the immediate rewards from the delayed rewards. The effectiveness of InferNet was evaluated on two online RL tasks: a simple GridWorld and 40 Atari games; and two offline RL tasks: GridWorld and a real-life Sepsis treatment task. For all tasks, the effectiveness of using the InferNet inferred rewards is compared against the immediate and the delayed rewards with two settings: with noisy rewards and without noise. Overall, our results show that the effectiveness of InferNet is robust against noisy reward functions and is an effective add-on mechanism for solving temporal CAP in a wide range of RL tasks, from classic RL simulation environments to a real-world RL problem and for both online and offline learning.
\end{abstract}


\section{Introduction}
\label{sec:delayed-reinforcement}
A large body of real-world tasks can be characterized as sequential multi-step learning problems, where the outcome of the selected actions is \emph{delayed}. Discovering which action(s) are responsible for the delayed outcome is known as the \textbf{\emph{(temporal) Credit Assignment Problem (CAP)}} \cite{Minsky1961StepsTA}.  
Solving the temporal CAP is especially important for \emph{delayed reinforcement} tasks \cite{sutton1985temporal}, in which a reward $r_t$ obtained at time $t$, can be affected by all previous actions, $a_0$, $a_1$, ..., $a_{t-1}$, $a_t$ and thus we need to assign credit or blame to each of those actions individually. Such tasks become extremely challenging if there are long delays between the actions and their corresponding outcomes.


Prior research has explored solving the CAP by formulating it as an RL problem \cite{suttonXreinforcement}, in which an agent learns how to interact with a potentially non-stationary, stochastic, and partially observable environment to maximize the long-term cumulative reward. For example, Temporal Difference (TD) learning methods \cite{sutton1988learning} have been widely used to tackle the CAP \cite{sutton1990time}. In particular, the TD($\lambda$) algorithm \cite{sutton1988learning,tesauro1992practical} uses eligibility traces to update the value of a state by using all the future rewards in the episode, which makes it easier to assign credit for long trajectories.

In prior work, one way to mitigate the impact of the CAP is to use model-based RL or simulations, which allow collecting vast amounts of data. However, in many real-life domains such as healthcare, building accurate simulations is especially challenging because disease progression is a rather complex process; moreover, learning policies while interacting with patients can be unethical or illegal. On the other hand, it is essential to solve the CAP problems in such domains because reward functions are often not only delayed but noisy. The most appropriate rewards to use in healthcare are the patient outcomes, which are typically unavailable until the entire trajectory is complete. This is because disease progression makes it difficult to assess patient health states moment by moment, and more importantly, many instructional or medical interventions that boost short-term performance may not be effective over the long term. Furthermore, reward functions in such domains are often incomplete or imperfect observations of the underlying true reward mechanisms. For example, a patient's final outcome of a stay can be inaccurate/noisy, as shown by the 30-day readmission rates among survivors of sepsis across 633,407 hospitalizations among 3,315 hospitals is 28.7\% \cite{sepsisarticle}. 

Previously, an alternative approach was proposed by \cite{azizsoltani2019unobserved}, denoted \emph{InferGP}. They first applied Gaussian Processes (GP) to infer \emph{unobservable} immediate rewards from delayed rewards, and then applied standard RL algorithms to induce policies based on the inferred rewards. While promising, that work has the following three limitations in the order of increasing severeness: 1) it did not investigate the effectiveness of InferGP with noisy reward functions, even though GP are known to be robust against noise; 2) InferGP does not scale up well, as it has poor time and space complexities, as shown below; and 3) InferGP can only be applied for offline-RL because it incorporates information from the entire training dataset into the model when applying Bayesian inference to infer the reward. Many DRL algorithms, however, often need millions or even billions of interactions obtained by extensively exploring the environment before they can learn a competitive policy, which makes InferGP impractical for large datasets and online RL.

In this work, we propose a novel Neural Network based approach named \emph{InferNet}, which infers ``immediate rewards" from the delayed rewards and then those inferred immediate rewards can be used to train any RL agent. InferNet is a general, \emph{scalable} mechanism that works alongside any online and offline RL algorithms. It is an easy yet effective add-on mechanism for mitigating the temporal CAP.  The effectiveness of InferNet was evaluated on two online RL tasks: a simple GridWorld and 40 Atari games; and two offline RL tasks: GridWorld and a real-life Sepsis treatment task. For both online and offline RL tasks, the effectiveness of using the InferNet inferred rewards is compared against immediate and delayed rewards. Additionally, we evaluated the effectiveness of each reward setting by adding noise to the reward functions to more accurately mimic real-world scenarios.

Our results shows that for online RL tasks, the InferNet policy significantly outperforms the delayed policy in the GridWorld, and its performance is on par with the immediate policy. For the Atari games, InferNet outperforms the corresponding delayed policy on 32 out of 40 games; and it can perform as well as or better than the immediate policy on 8 games. When noise is present in the rewards, InferNet outperforms the immediate policy on the GridWorld; and the performance of the immediate policies for the Atari games suffers greatly while the delayed and InferNet policies are less prone to be affected by the noise. Moreover, our proposed InferNet policy outperformed the corresponding delayed reward policy across 30 Atari games and it even performs better than or equal to the immediate reward policy on 23 games. On the offline RL tasks, InferNet performs as good as or better than both InferGP, immediate and delayed rewards. 

\section{InferNet: Neural Net Inferred Rewards}

\textbf{Problem Definition:} The environment is modeled as a Markov Decision Process, where at each time-step $t$ the agent observes the environment in state $s_t$, it takes an action $a_t$ and receives a scalar reward $r_t$ and the environment moves to state $s_{t+1}$. In the discrete case, $a_t$ is selected from a discrete set of actions $a_t \in A = \{1, ..., |A|\}$. The RL agent is tasked with maximizing the expected discounted sum of future rewards, or return, defined as $R_t = \sum_{\tau = t}^T \gamma^{\tau - t} r_t$, where $\gamma \in (0, 1]$ is the discount factor and $T$ is the last timestep in the episode. A value function is commonly used to estimate the expected return for each state or state-action pair. The optimal action-value function is defined as $Q^*(s, a) = max_\pi Q^\pi (s, a)$, where $Q^\pi$ estimates the long-term reward the agent would observe after following action $a$ from state $s$ and following policy $\pi$ thereafter.

\noindent \textbf{InferNet:} The intuition behind InferNet is rather straightforward. InferNet uses a deep neural network to infer the immediate rewards from the delayed reward in an episode. At each timestep, the observed state and action are passed as input to the neural network, which will output a single scalar, the inferred immediate reward for that state and action: $r_t = f(s_t, a_t | \theta)$. Here $\theta$ indicates the parameters (weights and biases) of the neural network. To address the credit assignment problem, InferNet distributes the final delayed reward among all the states in the episode. More specifically, the network learns to infer the immediate rewards from the delayed reward by applying a constraint on the predicted rewards: the sum of all the predicted rewards in one episode must be equal to the delayed reward, as shown in Equation \ref{inferred rewards} where $R_{del}$ indicates the delayed reward. This way, the network needs to model the reward function, conditioned on the state-action pair for each timestep, and it will minimize the loss between the sum of predicted rewards and the delayed reward for each episode. 
\begin{equation}
    \label{inferred rewards}
    R_{del} = f(s_0, a_0|\theta) + f(s_1, a_1|\theta) + ... + f(s_{T-1}, a_{T-1}|\theta)
\end{equation}
We used the \emph{TimeDistributed} layer available on TensorFlow Keras \cite{tensorflow2015-whitepaper,chollet2015keras} in order to repeat the same neural network operation multiple times, sharing weights across time, and pass the entire episode at once as input to the neural network. It should be noted that despite sharing weights across time, there is no internal state that is passed to the next timestep (as in a recurrent neural network). Each output is only dependent on the state and action passed as inputs at that timestep. We train InferNet by minimizing the loss function shown in Equation \ref{eq:loss function}. The pseudo-code for training InferNet alongside an RL algorithm is shown in Algorithm \ref{alg:main_alg}. This process can be seen as making the neural network learn a function that outputs a reward for each state-action pair, subject to the constraint of all rewards in an episode summing up to the delayed reward for that episode.
\begin{equation}
    \label{eq:loss function}
    Loss(\theta) = (R_{del} - \sum_{t=1}^{T} f(s_t, a_t|\theta) )^2
\end{equation}
To evaluate the effectiveness of InferNet, we divide the experimental evaluation into online and offline RL tasks.

\begin{algorithm}[tb]
\caption{InferNet Online}
\label{alg:main_alg}
\begin{algorithmic}[1] 
\STATE Initialize InferNet buffer $D \leftarrow ()$
\STATE // Pretrain InferNet
\FOR{$episode \gets1$ to $K$}
    \STATE Play an episode randomly and collect the data
    \STATE Delayed reward $R_{del} = r_0 + r_1 + .. + r_{T-1}$
    \STATE $D \leftarrow D \cup (s_0, a_0, ..., s_{T-1}, a_{T-1}, R_{del})$
    \STATE Sample mini-batch of episodes $B \sim D$
    \STATE Train InferNet on $B$: \\ $L(\theta) = (R - \sum_{t=0}^{T-1} f(s_t, a_t)|\theta))^2$
\ENDFOR
\FOR{$episode \gets1$ to $M$}
    \STATE    Set episode data sequence $tmp \leftarrow ()$
    \WHILE{not end of episode}
        \STATE Get state $s$ from env
        \STATE Select action $a \sim \pi$
        \STATE $s', r \sim env(s, a)$
        \STATE $tmp \leftarrow tmp \cup (s, a, r, s')$ 
        \STATE Train RL agent
        \STATE Sample batch of episodes $B \sim D$
        \STATE Train InferNet on $B$: \\ $L(\theta) = (R - \sum_{t=0}^{T-1} f(s_t, a_t)|\theta))^2$
    \ENDWHILE
    \STATE Use InferNet to infer rewards for the steps in $tmp$\\
    \STATE Replace rewards in $tmp$ with InferNet rewards\\
    \STATE $D \leftarrow D \cup tmp$\\
    \STATE Store data in $tmp$ to train the RL agent later on
\ENDFOR
\end{algorithmic}
\end{algorithm}

\section{Online RL Experiments}

The effectiveness of InferNet is investigated on a GridWorld first, and then on the Atari 2600 Learning Environment. We compare the following reward settings: 1) \emph{Immediate rewards:} when available, they are the gold standard. 2) \emph{Delayed rewards:} these rewards are used as a baseline. All the intermediate rewards will be zero and the reward that indicates how good or bad the intermediate actions were will only be provided at the end of the episode. When the rewards are not delayed by nature, we simulate the delayed rewards by ``hiding" the immediate rewards and providing the sum of all the immediate rewards at the end of the episode, as one big delayed reward. 3) \emph{InferNet rewards:} our proposed method which uses a Neural Network to predict the immediate rewards from the delayed reward. On both tasks, we also evaluate the power of InferNet when the reward function is noisy.

\subsection{Grid World}
\label{sec:gridworld}
A GridWorld environment was designed as a simple RL testbed where we can compare the true immediate rewards to the inferred rewards produced by InferNet. This environment consists of a 14x7 grid, with five positive rewards (+1) and four negative rewards (-1), located randomly along the grid, but always in the same locations. All other states have a reward of zero. The initial state is located at the bottom-right corner of the grid, and the agent's goal is to reach the terminal state, located at the top-left corner, while collecting the positive rewards and avoiding the negative ones. The three available actions are: move up, left and down.

Figure \ref{fig:losses} (Appendix) shows that by minimizing the training error in Eq. \ref{eq:loss function} (the difference between the delayed reward and the sum of immediate predicted rewards) (red line), InferNet minimizes the true error (the difference between the predicted and true immediate rewards) (blue line). We then evaluate the effectiveness of the rewards predicted by InferNet when used to train an RL agent, and compare them to the immediate and delayed rewards. We repeated each experiment five times with different random seeds, and show the mean and standard deviation of those runs. We explored Q-Learning and TD($\lambda$) as RL agents; the latter is known for being able to solve the temporal CAP.

\begin{figure}[t]
    \centering
    \includegraphics[width=0.34\columnwidth]{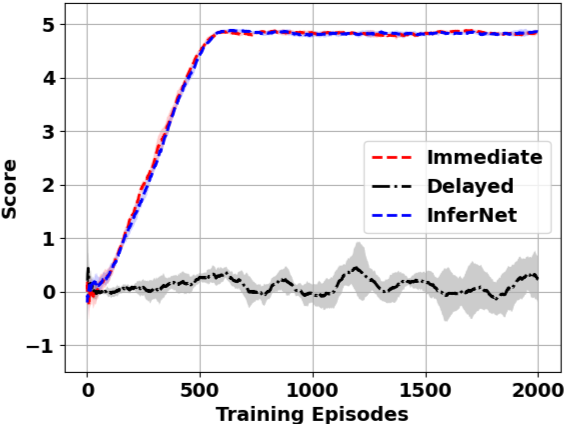}
    \includegraphics[width=0.32\columnwidth]{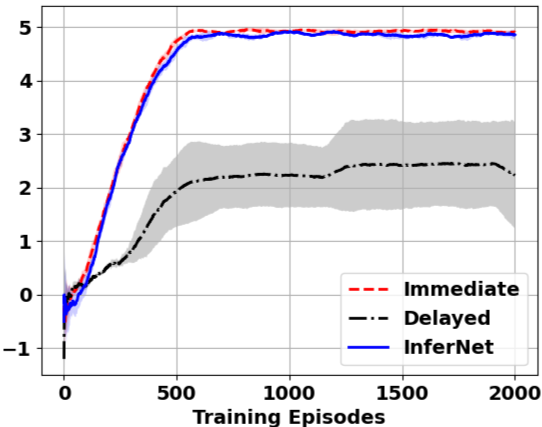}
    \includegraphics[width=0.34\columnwidth]{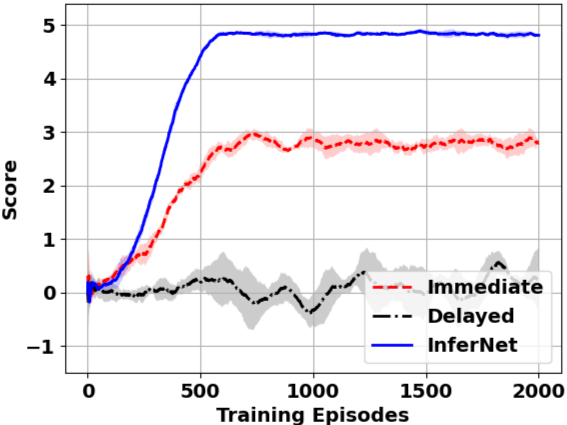}
    \includegraphics[width=0.32\columnwidth]{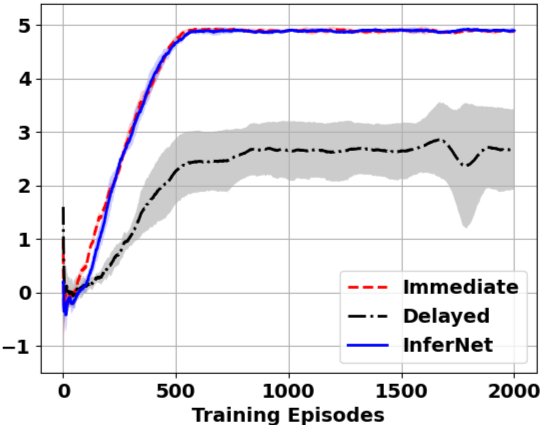}
    \caption{Results of training a Q-Learning agent (left) and a TD($\lambda$) agent (right) on the GridWorld task with no noise, for immediate, delayed, and InferNet rewards. No noise in the rewards (Top) and noisy rewards (Bottom). Gaussian noise ($\mathcal{N}(0, 0.08^2)$)}
    \label{fig:gridworld-online-results}
\end{figure}

\noindent \textbf{Q-Learning }
is a TD-Learning method, but it does not employ any eligibility traces, and it usually only uses a 1-step reward. Figure \ref{fig:gridworld-online-results} (Top Left) shows the result of training different Q-Learning agents on the three reward settings without noise. These results clearly show that the Q-Learning agent that uses the delayed rewards cannot learn to solve the simple GridWorld task. It is no better than a random agent. However, when we first use InferNet to infer the ``immediate" rewards from the delayed ones, the agent is able to solve the task as effectively as the agent uses the immediate rewards. When the reward function is noisy, Figure \ref{fig:gridworld-online-results} (Bottom Left) shows that the immediate reward agent suffers significantly, and cannot solve the environment completely, while InferNet is much more robust to the noise.

\noindent \textbf{TD($\lambda$) }
is known to be one of the strongest methods to solve the CAP. This algorithm takes advantage of the benefits of TD methods, and includes eligibility traces, which allows the agent to look at all the future rewards to estimate the value of each state. This makes propagating the delayed reward easier than in the case of 1-step rewards. Despite all these advantages, Figure \ref{fig:gridworld-online-results} (Top Right) shows that when the rewards are delayed, the agent is not able to learn as effectively as the agent that has access to the true immediate rewards. However, the agent that uses the InferNet predicted reward achieves the same performance as the agent that uses the immediate rewards; they can both fully solve the environment. When the reward function is noisy, Figure \ref{fig:gridworld-online-results} (Bottom Right) shows that none of the agents suffer. This result shows that TD($\lambda$) is more robust to noisy rewards than Q-Learning.

\subsection{Atari Learning Environment (ALE)}
The ALE provides visually complex environments in which the state space is very high dimensional, represented by pixels on a screen. It is important to note that in some games, each episode can consist of thousands of steps, so learning from a single delayed reward is no trivial task. We used OpenAI gym \cite{1606.01540} to simulate the environments, and the stable baselines library \cite{stable-baselines} to train the Deep RL agent. Here we evaluate the performance of InferNet in conjunction with a Prioritized Dueling DQN agent \cite{wang2015dueling,Schaul2015_priortized}.

\begin{figure*}[t]
    \begin{center}
        \includegraphics[width=0.42\columnwidth]{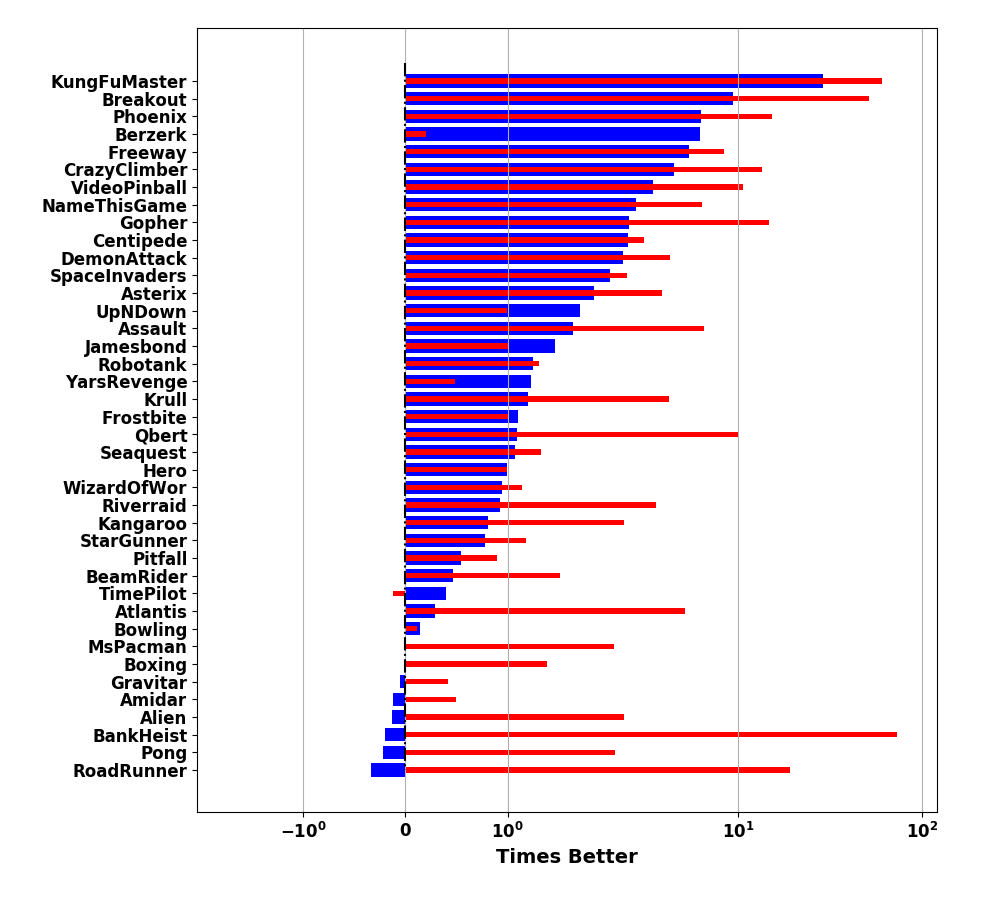}
        \includegraphics[width=0.42\columnwidth]{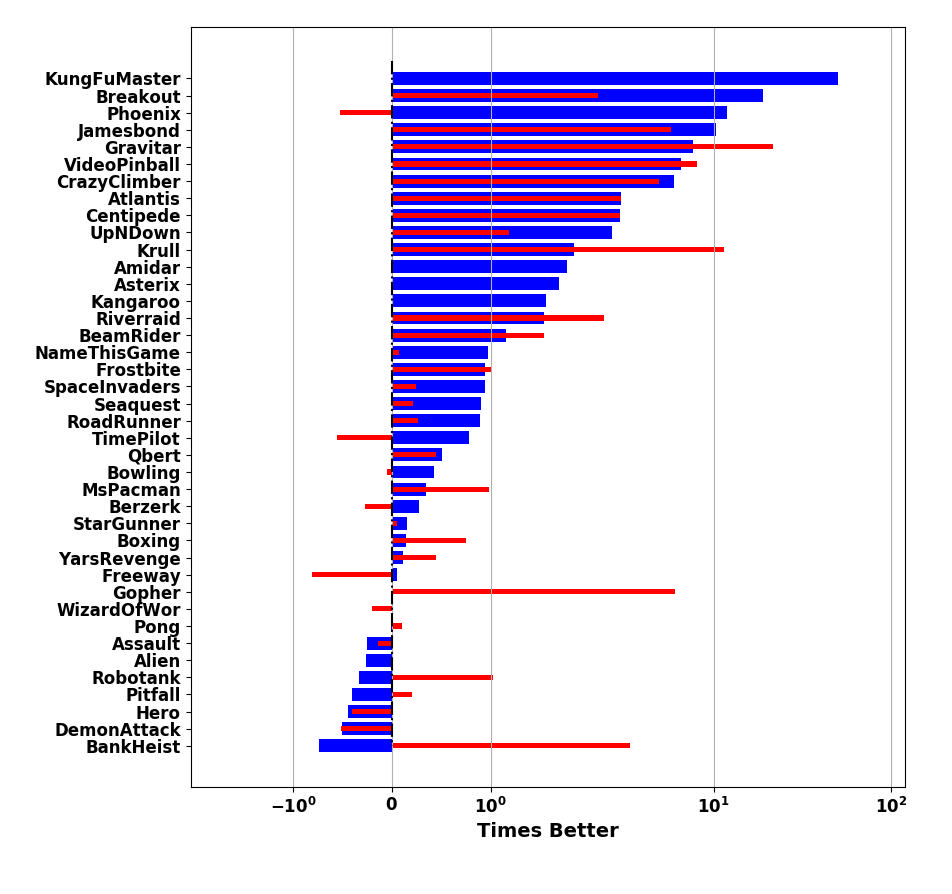}
        \caption{Performance of the Prioritized Dueling DQN agent on the different Atari games with the three reward settings: Delayed (vertical line at x=0), Immediate (red) and InferNet (Blue). Without noisy rewards (Left) and with noisy rewards (Right). The results have been normalized to show the Delayed rewards at x=0 as a vertical line, and each bar shows how many times better than the delayed agent it is.}
        \label{fig:Atari}
    \end{center}
\end{figure*}

\noindent \textbf{Noise-free Rewards:}
The results of training the Dueling DQN agent on the three different reward settings are shown in Figure \ref{fig:Atari} (Left). The full results can be found in the supplementary material. The agent trained on the rewards provided by InferNet performs as well as or better than the agent which uses the delayed rewards in almost all games. In some cases, it can even match the performance of the agent in the Immediate setting. These results clearly show that when immediate rewards are not available, using our InferNet is preferable over training the agent on the delayed rewards.

\begin{figure}[t]
    \centering
    \includegraphics[width=0.37\columnwidth]{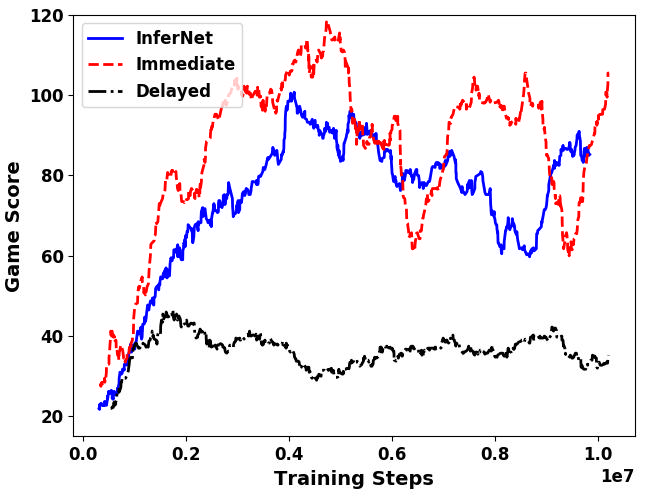}
    \includegraphics[width=0.37\columnwidth]{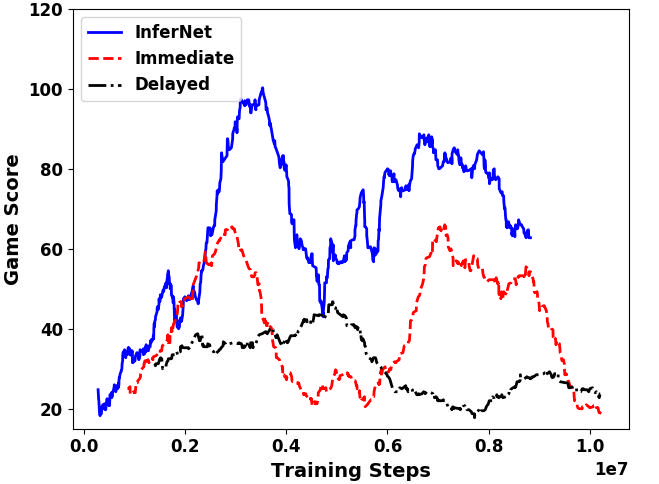}
    \includegraphics[width=0.37\columnwidth]{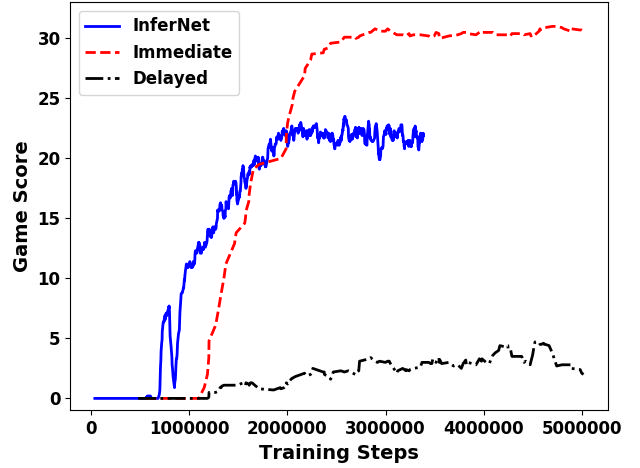}
    \includegraphics[width=0.37\columnwidth]{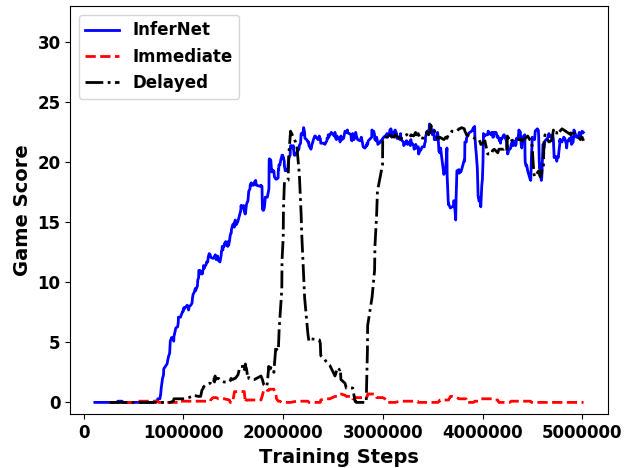}
    \caption{Training process for Seaquest (Top) and Freeway (Bottom). Left: No Noise. Right:  Gaussian noise ($\mathcal{N}(0, 0.6^2)$).}
    \label{fig:seaquest-freeway}
\end{figure}

\noindent \textbf{Noisy Rewards:}
We repeated the Atari experiments after adding Gaussian noise to the observed rewards. As the noise is unbiased, the expectation of the sum of rewards is the same with and without noise, as shown in Eq. \ref{noise info}.
\begin{dmath}
    \label{noise info}
    \mathbb{E}[R] = \mathbb{E}[r_0 + ... + r_{T-1}] = \mathbb{E}[r_0 + \mathcal{N}(0, \sigma^2) + ... + r_{T-1}+ \mathcal{N}(0, \sigma^2)]
\end{dmath}

Figure \ref{fig:Atari} (Right) shows the results of training the same Prioritized Dueling DQN agent on noisy rewards on immediate, delayed and InferNet rewards. It shows that the performance of the agent trained on noisy immediate rewards suffers significantly when compared to the noisy-free immediate rewards. InferNet outperforms the Immediate rewards in more games than in the noise-free setting. Two examples of this are the games of Seaquest and Freeway (Figure \ref{fig:seaquest-freeway}).

\section{Offline RL Experiments}
When applying RL to solve many real-life tasks such as healthcare, we have to perform offline learning. This means that the training needs to be done from a fixed training dataset, and no further exploration of the environment is possible. In these situations, having a method that effectively solves the CAP is crucial. For the offline experiments, we added one more reward setting to the experiment: InferGP. This is an alternative prior method for inferring the immediate rewards from the delayed ones. Prior work has shown that InferGP works reasonably well in a wide range of offline RL tasks \cite{azizsoltani2019unobserved}. Our goal is to determine the efficacy of InferNet for offline RL tasks, when compared to immediate, delayed and InferGP rewards. We use the same GridWorld environment as in Section \ref{sec:gridworld}. Additionally, we want to evaluate our method in a real world problem: a healthcare task where the goal of the agent is to induce a policy for sepsis treatment and septic shock prevention.

\begin{figure}[t]
    \centering
    \includegraphics[width=0.495\columnwidth]{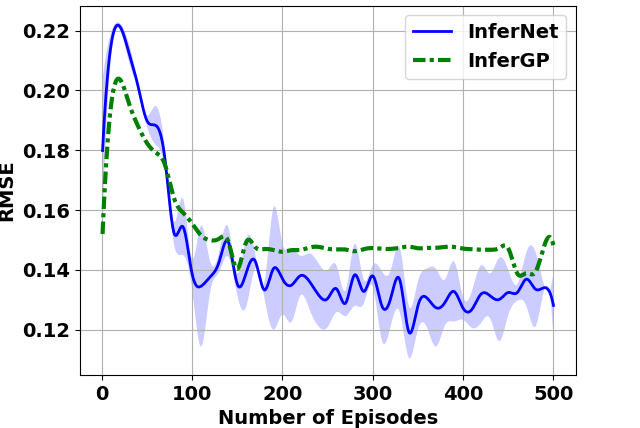}
    \includegraphics[width=0.495\columnwidth]{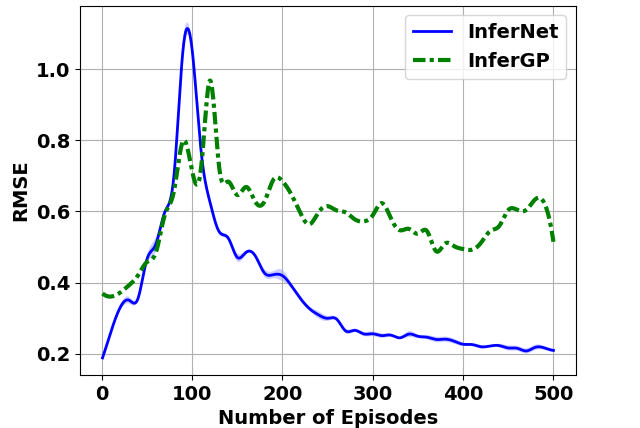}
    \caption{RMSE between the inferred and the true immediate rewards as a function of the number of episodes collected from the GridWorld environment. Left: No noise. Right: Gaussian noise ($\mathcal{N}(0, 0.3^2)$).}
    \label{fig:rmse-gridworld}
\end{figure}

\subsection{GridWorld}
In this offline experiment, we generate random data, and then use that data to infer the rewards and train the RL agent.

\noindent \textbf{RMSE:}
We evaluated the amount of data needed for InferNet and InferGP to approximate the true immediate rewards. We calculated the root mean squared error (RMSE) between the inferred rewards and the true immediate rewards in the training dataset by varying the amount of training data. Figure \ref{fig:rmse-gridworld} (Left) compares this RMSE for InferNet and InferGP. Overall, with 100 or more trajectories, InferNet consistently has a lower RMSE than InferGP. Figure \ref{fig:rmse-gridworld} (Right) compares the RMSE of the two approaches with noisy rewards. Adding noise makes the CAP more challenging, and InferGP cannot adapt as well as InferNet (InferGP increases from 0.15 to 0.5 and InferNet increases from 0.13 to 0.2), despite the fact that GP are known to be able to handle noisy data.

\begin{figure}[t]
    \centering
        \centering
        \includegraphics[width=0.495\columnwidth]{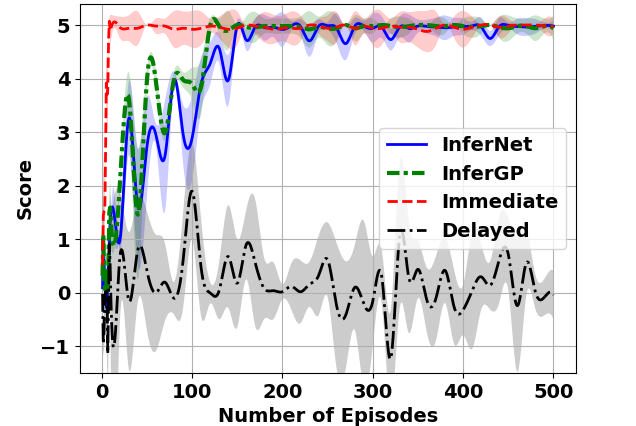}
        \centering
        \includegraphics[width=0.495\columnwidth]{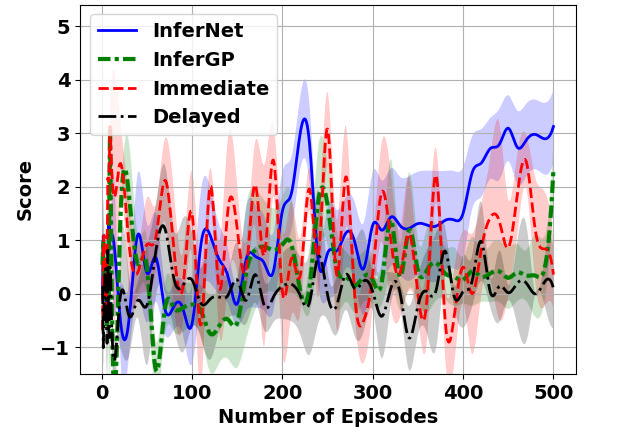}
    \caption{Performance of Q-Learning agents on the GridWorld environment as a function of the number of training episodes. Left: No noise. Right: Gaussian Noise ($\mathcal{N}(0, 0.3^2)$)}
    \label{fig:offline-gridworld}
\end{figure}

\noindent \textbf{Offline Q-Learning:}
We trained a tabular Q-learning agent for 5000 iterations on the same dataset used to infer the rewards. We compared the four reward settings: immediate, delayed, InferGP and InferNet. Once our RL policies are trained, their effectiveness is evaluated online by interacting with the GridWorld environment directly. Figure \ref{fig:offline-gridworld} (Left) shows the mean and standard deviation of the performance of the agent when interacting with the environment for 50 episodes, as a function of the number of episodes available in the training dataset. Figure \ref{fig:offline-gridworld} (Left) shows that, as expected, the delayed policy performs poorly, while the Immediate policy can converge to the optimal policy after only 10 episodes of data; additionally, both InferNet and InferGP both can converge to the optimal policy but they need more training data (around 150 episodes) than the Immediate policy. Figure \ref{fig:offline-gridworld} (Right) shows the performance of the policies when the rewards are noisy. It clearly shows that adding noise to the rewards function deteriorates the Immediate policy, while InferNet and InferGP policies also suffer but it seems that InferNet is the best option.

\begin{wrapfigure}{r}{0.4\columnwidth}
    \centering
    \includegraphics[width=0.49\columnwidth]{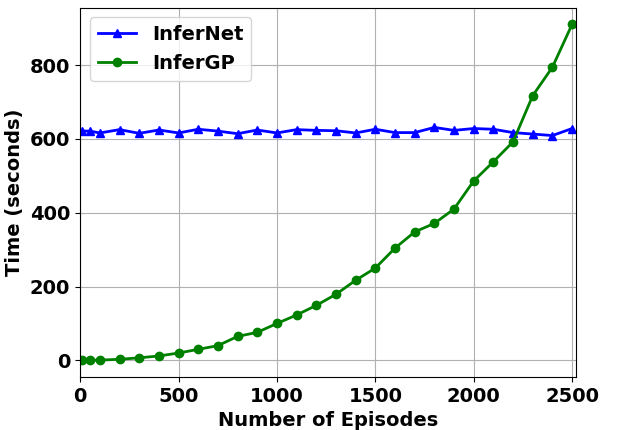}
    \caption{Time analysis of InferNet and InferGP.}
    \label{fig:time-gridworld}
\end{wrapfigure}

\noindent \textbf{Time Complexity}: Figure \ref{fig:time-gridworld} empirically compares the time complexity of InferNet and InferGP: the training time of InferNet is less sensitive to the size of the training dataset, while the training time of InferGP increases cubically as the training data increases. Fundamentally, InferGP has an asymptotic time complexity of $O(n^3)$ and an asymptotic space complexity of $O(n^2)$, where $n$ refers to the size of the dataset. InferNet has a time complexity of $O(n)$ since we sample a constant amount of mini-batches from the dataset for each gradient descent step, and we only need to train the network for a constant number of epochs. The space complexity for InferNet is $O(f*l)$, where $f$ is the number features in the state and action that are passed as inputs, and $l$ is the length of the episode that is passed as input.

\subsection{Healthcare}



We evaluate InferNet and InferGP, on a real-world sepsis treatment task where the rewards are delayed. The goal is to learn an optimal treatment policy to prevent patients from going into septic shock, the most severe complication of sepsis, which leads to a mortality rate as high as 50\%. As many as 80\% of sepsis deaths could be prevented with timely diagnosis and treatment \cite{kumar2006duration}; thus, it is crucial to monitor sepsis progression and recommend the optimal treatment as early as possible. Despite the severity of the disease and the challenges faced by practitioners, it is notoriously difficult to reach an agreement for the optimal treatment due to the complex nature of sepsis and different patients’ constitutions. Moreover, continuous updates in the sepsis guidelines often lead to inconsistent clinical practices \cite{Backer2017}. Recently, several DRL approaches have been investigated for septic treatment, utilizing Electronic Health Records (EHRs). However, \cite{Raghu2017,Komorowski2018} only considered delayed rewards, while \cite{azizsoltani2019unobserved} leveraged the Gaussian process based immediate reward inference method, which is one of our baselines. 


\noindent \textbf{Data:} Our EHRs were collected from a large US healthcare system (July, 2013 to December, 2015). We identified $2,964$ septic shock positive visits and sampled $2,964$ negative visits based on the expert clinical rules, keeping the same ratio of age, gender, race, and the length of hospital stay as in the original EHR. We selected 22 sepsis-related state features such as vital signs, lab results and medical interventions, and defined four types of treatments as actions: no treatment, oxygen control, anti-infection drug, and vassopressor.

\noindent \textbf{Reward:} 
The rewards were assigned by the expert-guided reward function based on the multiple septic stages. These rewards are delayed in time and noisy. The delayed rewards are given when the patient goes into septic shock or recovers at the end of their stay, and the noise in the rewards is a result of imperfect sensors or incomplete measurements.

\noindent \textbf{Experiment setting:} We compared three reward settings: delayed, InferGP, and InferNet, since the immediate rewards are not available. InferNet predicts one reward for each timestep. After inferring the immediate rewards, the septic treatment policies were induced using a Dueling DQN agent. The hyper-parameters are shown in the Appendix.


\noindent \textbf{Evaluation metric:} The induced policies were evaluated using the \emph{septic shock rate}, which is the portion of shock-positive visits in each group belonging to the corresponding agreement rate, as the agreement rate increases from 0 to 1 with a 0.1-rate interval. It is desirable that the higher the agreement rate, the lower the septic shock rate. The policy agreement rate in the non-shock patients should be higher, which means that our policy agrees more with the physicians' actions for the non-shock patients.

\begin{figure}[t]
    \centering
    \includegraphics[width=0.48\columnwidth]{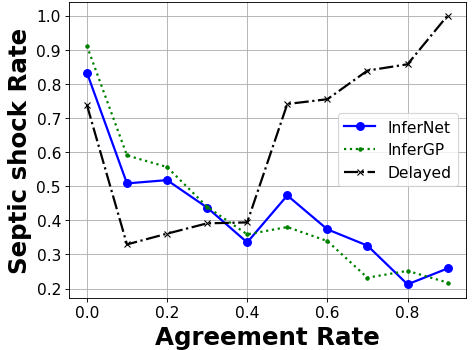}
    \includegraphics[width=0.48\columnwidth]{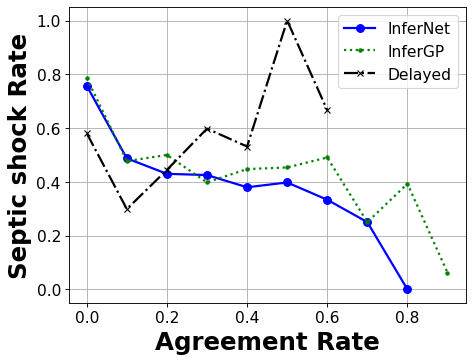}
    \caption{Healthcare: Septic shock rate as a function of the agreement rate between the policy and the physician actions for the training (Left) and test (Right) sets.}
    \label{fig:ehr}
\end{figure} 

\noindent \textbf{Results:} The underlying assumption is that the agreed treatments are adequate and acceptable, as they were taken by real doctors in real clinical cases, but not necessarily optimal. Figure \ref{fig:ehr} shows the \emph{septic shock rate} in the visit group for the training (Left) and test (Right) sets, as a function of the corresponding agreement rates. For InferNet, the septic shock rate almost monotonically decreases in the test set evaluation, as the agreement rate increases and reaches the lowest shock rate of all policies. InferGP shows a general trend of decreasing shock rate with a larger variance than InferNet as the agreement rate increases, while delayed fails to learn an effective shock prevention policy. This supports that InferNet significantly improves the policy training process at preventing septic shock, compared with InferGP and delayed.
Furthermore, InferNet and InferGP induced policies that agree with the physicians more for the non-shock patients, while when the patients are more likely to go into septic shock, the agents try to search a different treatment strategy from the given treatments that resulted in septic shock.

\section{Related Work}
In recent years, by utilizing deep learning and novel RL algorithms, Deep RL (DRL) has shown great success in various complex tasks \cite{berner2019dota,silver2018general}. Much of prior work on DRL has focused on online learning where the agent learns while interacting with the environment. Immediate rewards are generally much more effective than delayed rewards for RL because of the CAP: the more we delay rewards or punishments, the harder it becomes to assign credit or blame properly. Different approaches have been proposed and applied for solving the CAP. For example, when applying DRL for games such as Chess, Shogi and Go, the final rewards are determined by the outcomes of the game: $-1$/$0$/$+1$ for loss/draw/win respectively; and for each state, Monte Carlo Tree Search (MCTS) was used to learn the likelihood of each outcome \cite{silver2017mastering,silver2018general}. Because of the CAP, RL algorithms often need more training data to learn an effective policy using delayed rewards than using immediate rewards. More importantly, for some extremely complicated games, DRL may fail to learn an effective policy altogether. As a result, prior research used expert-designed immediate rewards, or learned a reward function from expert experience trajectories, using reward engineering methods such as Inverse RL \cite{ziebart2008maximum,abbeel2004apprenticeship,levine2011nonlinear,ramachandran2007bayesian}. For example, Berner et al. used human-crafted intermediate rewards to simplify the CAP. They designed a reward function based on what expert players agree to be good in that game \cite{berner2019dota}. While effective, such expert-designed rewards are often labor-intensive, expensive, and domain specific. Additionally, these expert rewards might introduce expert bias into the process, leading to sub-optimal agent performance, as shown by AlphaGo Zero \cite{silver2017mastering} outperforming the original AlphaGo \cite{silver2016mastering}.

The human brain is very efficient at solving the CAP when learning to perform new tasks \cite{asaad2017prefrontal,richards2019dendritic}. Thus a wealth of neuroscience research focuses on understanding the learning and decision-making process in animals and humans. For example, \cite{agogino2004unifying} studied the structural and temporal CAP and suggested a unification of the problem for multi-agent, time-extended problems. In RL, the temporal CAP has been widely studied \cite{sutton1985temporal}, and solutions to it have been proposed in order to more successfully train neural network systems \cite{ororbia2018conducting,lansdell2019learning}.

In machine learning, prior research tried to solve the CAP by formulating it as an RL task \cite{suttonXreinforcement}. The best known family of algorithms to tackle the CAP are the Temporal Difference (TD) Learning methods, and TD($\lambda$) in particular \cite{sutton1988learning}. It employs eligibility traces to use all the future rewards when updating the value of each state, resulting in better assignment of credit/blame for each action.

\section{Conclusion}
We developed a deep learning algorithm that explicitly tackles the CAP by generating immediate rewards from the delayed rewards. Our results show that our algorithm makes it easier for the RL algorithm of choice to solve the task at hand, both for online and offline RL, while mitigating the problem generated by noisy rewards. We showed that InferNet can accurately predict the true immediate rewards on a simple GridWorld and help Q-Learning and TD($\lambda$) agents solve the environment. An RL agent that learns a treatment to avoid septic shock from a real-life healthcare dataset can also benefit from the rewards provided by InferNet to make more effective decisions. Finally, we showed that our algorithm scales to large datasets and to online RL, which allows it to help solve more complex pixel-based games such as the Atari games, and it can be especially useful when the reward is noisy as shown by the performance of the agent on the noisy version of the Atari games.


\bibliographystyle{unsrt}
\bibliography{paper}

\newpage

\appendix

\section{Additional Results}

\subsection{GridWorld}
\begin{figure}[t]
    \centering
    \includegraphics[width=0.4\columnwidth]{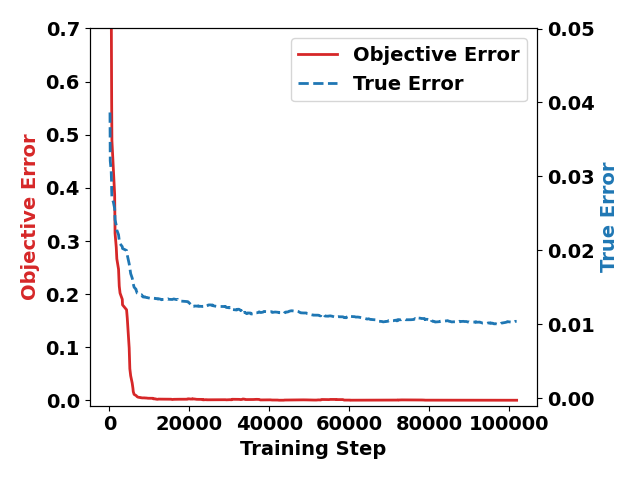}
    \caption{Training process for InferNet. Minimizing the objective loss (red line) results in also minimizing the true loss (blue line), which is what we want to ultimately achieve.}
    \label{fig:losses}
\end{figure}

As mentioned in section \ref{sec:gridworld}, Figure \ref{fig:losses} shows that by minimizing the training error in Eq. \ref{eq:loss function} (the difference between the delayed reward and the sum of immediate predicted rewards) (red line), InferNet minimizes the true error (the difference between the predicted and true immediate rewards) (blue line). This result empirically shows that minimizing our objective loss function is indeed making InferNet learn the true immediate rewards.

\subsection{CartPole}
The goal in CartPole is to move a cart left and right in order to keep a pole balanced in its vertical position. The reward function provides a rewards of +1 for each step where the pole is kept vertical. This means that the reward function InferNet needs to learn is pretty simple, a reward of +1 for each timestep, regardless of the state and action passed as inputs. The difficulty in this environment lies on the continuous state space, that makes tabular RL methods inoperable. For this reason, we compared the same three reward settings using the DQN algorithm. For the InferNet setting, we trained the DQN agent \cite{mnih2015human} and InferNet simultaneously, providing the DQN agent with the rewards produced by InferNet.

\begin{figure}[t]
    \centering
    \includegraphics[width=0.4\columnwidth]{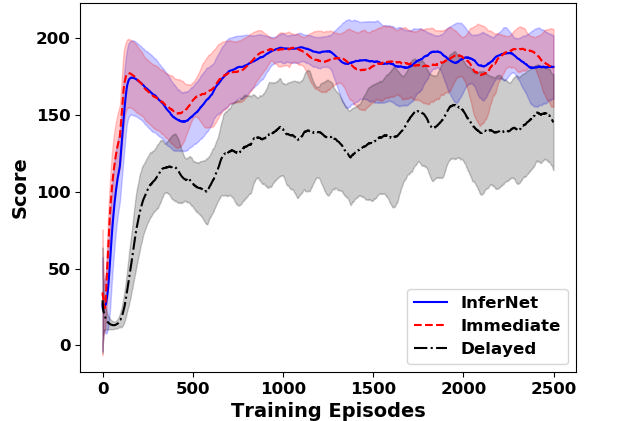}
    \caption{Comparison of immediate, delayed and InferNet policies in the CartPole task.}
    \label{fig:cartpole}
\end{figure}

The results of training this DQN agent on the immediate, delayed and InferNet rewards are shown in figure \ref{fig:cartpole}, where we show the mean and standard deviation of training each agent with 20 different random seeds as a function of the number of training steps. These results show that InferNet can work in combination with a Deep RL agent, to mitigate the credit assignment problem and boost the performance of the agent when only delayed rewards are available, and can perform as well as the agent that uses the immediate rewards. Meanwhile, the agent that learns from the delayed rewards cannot completely solve this task, although it is able to get a reasonable score.

\subsection{Impact of Different Noise Levels}

\begin{figure*}[t!]
    \centering
    \includegraphics[width=0.32\columnwidth]{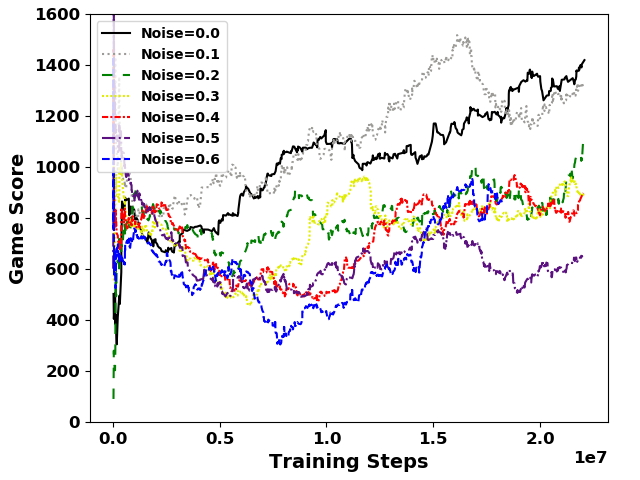}
    \includegraphics[width=0.32\columnwidth]{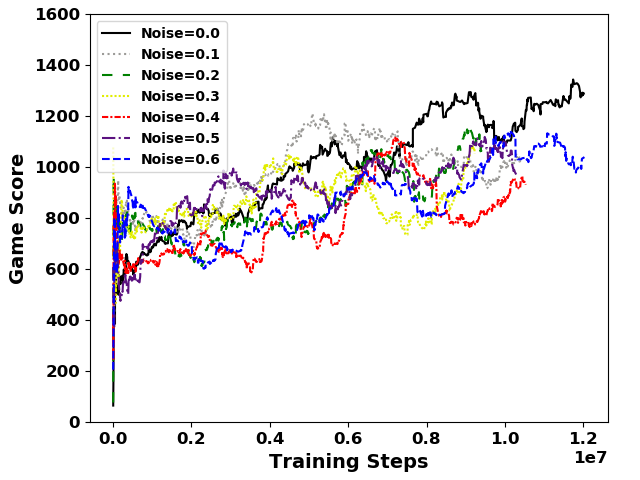}
    \includegraphics[width=0.32\columnwidth]{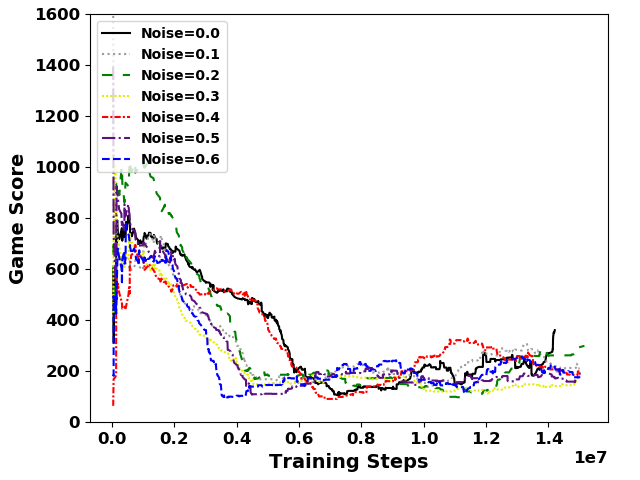}
    \caption{Training on Centipede with different noise levels. Left: Immediate. Center: InferNet. Right: Delayed.}
    \label{fig:centipede}
\end{figure*}

\begin{figure*}[h]
    \centering
    \includegraphics[width=0.32\columnwidth]{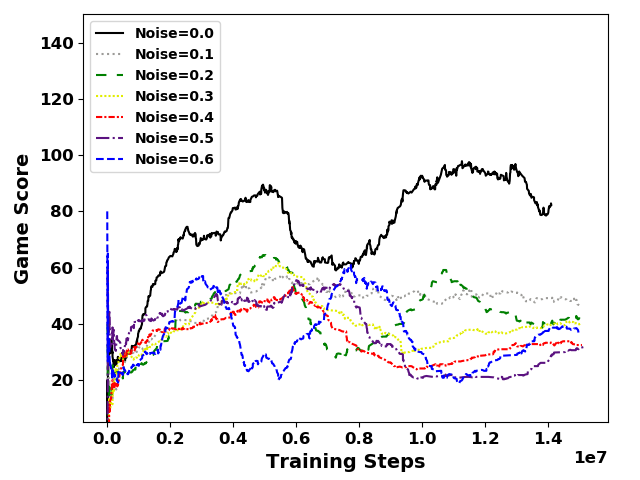}
    \includegraphics[width=0.32\columnwidth]{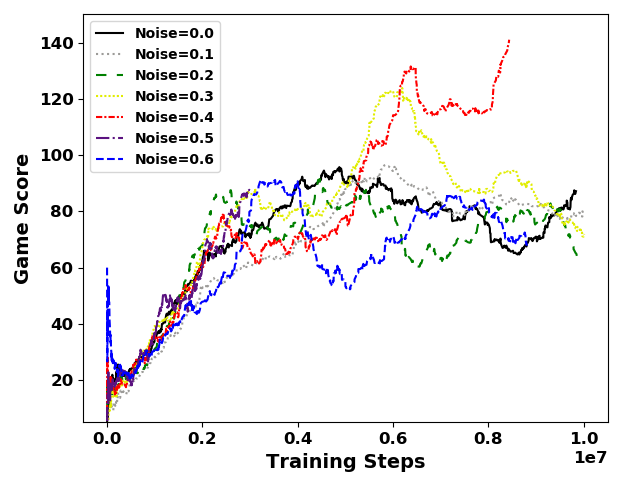}
    \includegraphics[width=0.32\columnwidth]{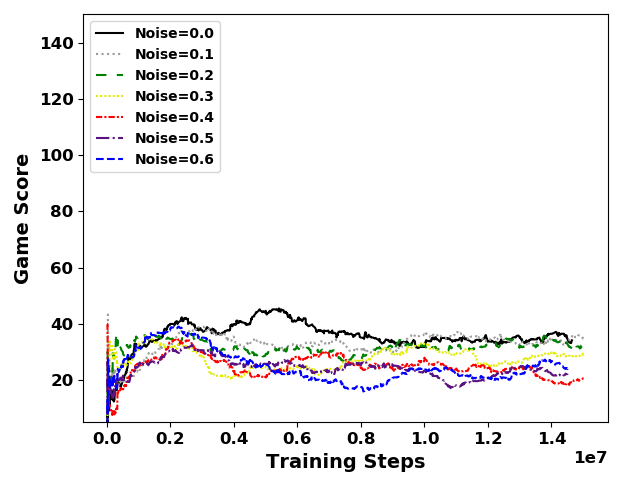}
    \caption{Training on Seaquest with different noise levels. Left: Immediate. Center: InferNet. Right: Delayed.}
    \label{fig:seaquest}
\end{figure*}

\begin{figure*}[t!]
    \centering
    \includegraphics[width=0.32\columnwidth]{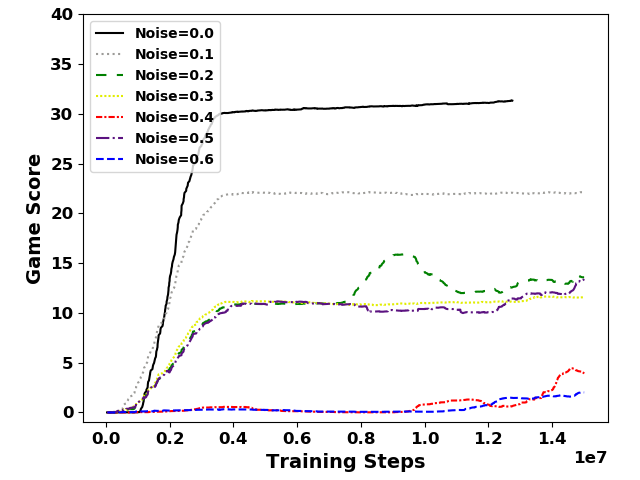}
    \includegraphics[width=0.32\columnwidth]{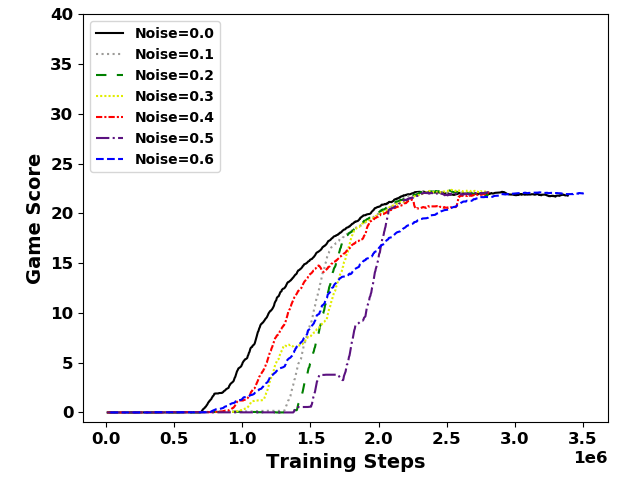}
    \includegraphics[width=0.32\columnwidth]{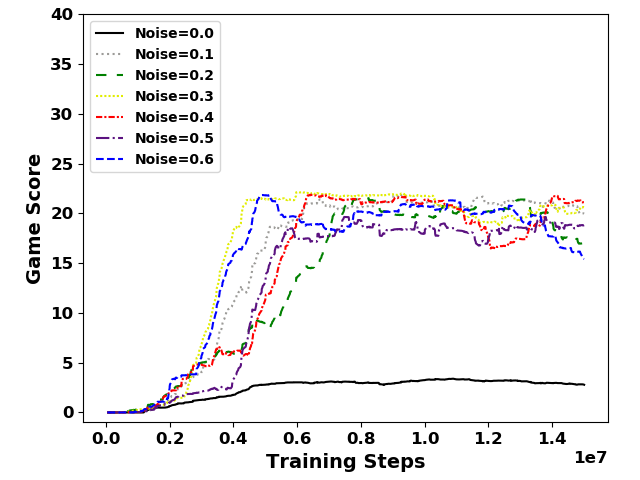}
    \caption{Training on Freeway with different noise levels. Left: Immediate. Center: InferNet. Right: Delayed.}
    \label{fig:freeway}
\end{figure*}

In this section of the appendix, we explore the impact that different levels of noise have on the different reward settings. For that, we chose three Atari games where the Immediate reward setting was impacted in several ways: Centipede, Freeway and Seaquest. The standard deviation of the Gaussian distribution used to generate the white noise (the mean is still zero) was modified to adjust the overall level of noise. We tried the following standard deviations: 0.0, 0.1, 0.2, 0.3, 0.4, 0.5, 0.6. Figures \ref{fig:centipede}, \ref{fig:seaquest}, and \ref{fig:freeway} show the mean performance of the agent over two runs with different random seeds. They show how the different noise levels affect the performance of each setting for these three games. InferNet is probably the method that best balances robustness to noise with overall performance.

\newpage 

\subsection{Atari Learning Environment}
The NN architecture for Deep RL agent was the same as in \cite{wang2015dueling}). InferNet also uses the same architecture as the agent. The only differences were 1) the output layer consists of a single value for InferNet, 2) InferNet uses dropout during training for regularization, and 3) we wrapped all the InferNet layers in the TimeDistributed layer, in order to be able to pass all the steps in an episode as input at once, and infer the immediate reward for each of those steps (see Algorithm \ref{alg:main_alg}). It should be noted that the hyper-parameters (including the replay buffer size) are different from those in the Dueling DQN paper. We allowed for these changes because our goal is not to outperform the current state of the art method or to improve upon some prior Deep Reinforcement Learning algorithm. Our task is to create a method that infers better rewards and can be used by other RL algorithms in order to learn more effectively. Modifying some of these hyper-parameters allowed for less expensive training.

Table \ref{final-table} shows the complete results of the Prioritized Dueling DQN agent, with the different reward settings, on all the Atari games. However, it is important to know that as we did not intend to outperform any other algorithm or state-of-the-art method, our evaluation was different from the Dueling DQN paper in several ways: 1) the size of the experience replay buffer was 10,000 instead of 1,000,000; 2) the evaluation was performed by making the loss of a life indicate the end of the episode; 3) we used the default hyper-parameters in the stable baselines library (which can be seen in our source code too).

We also show the approximate episode length to get an idea of the situations in which InferNet may perform best. Finally, we moved some of the games to the lower section of the table for three possible reasons, 1) none of the agents learned anything useful, they are not better than random (shown in the lower section of the table, but the different columns do include a number); 2) the need for more computational resources needed to train InferNet on these games is too large (shown in the lower section of the table, and the columns are left blank); 3) other runtime errors (shown in the lower section of the table, and the columns are left blank).

\begin{table*}[b!]
\caption{Performance of the different settings on the Atari games, starting with 30 no-op actions. The evaluation was different from the usual in: 1) the size of the experience replay buffer was 10,000 instead of 1,000,000; 2) the evaluation was performed by making the loss of a life indicate the end of the episode; 3) we used the default hyper-parameters in the stable baselines library.}
\label{final-table}
\begin{center}
\begin{small}
\begin{sc}
\begin{tabular}{|l|rrr|rrr|r|}
\toprule
 & & Non-Noisy & &                                              & Noisy & & Approx. Ep. Length \\
\midrule
Game                & Immediate & InferNet & Delayed            & Immediate & InferNet & Delayed & \\
\midrule
Amidar              & \textbf{24} & 14 & 16                      & 5.7 & \textbf{15.8} & 5.7 & 200 \\
Seaquest            & \textbf{93} & 83 & 40                      & 51 & \textbf{80} & 42 & 200 \\
Space Invaders      & \textbf{202} & 175 & 58                    & 83 & \textbf{129} & 66.5 & 220 \\
Star Gunner         & \textbf{240} & 196 & 110                   & 82 & \textbf{90} & 78 & 220 \\
Wizard Of Wor       & \textbf{75} & 68 & 35                      & 30 & \textbf{45} & 25 & 170 \\
Asterix             & \textbf{970} & 570 & 200                   & 75 & \textbf{202} & 75 & 200 \\
Battle Zone         & \textbf{1000} & 955 & 885                  & 982 & \textbf{1043} & 730 & 300 \\
Breakout            & \textbf{42} & 8.3 & 0.8                    & 2.6 & \textbf{16} & 0.8 & 120 \\
Crazy Climber       & \textbf{10,070} & 3850 & 700               & 3560 & \textbf{4200} & 600 & 800 \\
Freeway             & \textbf{32} & 21.7 & 3.4                   & 4 & \textbf{21.7} & 20.7 & 2000 \\
Kangaroo            & \textbf{715} & 380 & 210                   & 48 & \textbf{123} & 48 & 180 \\
Kung-Fu Master      & \textbf{2765} & 1350 & 45                  & 30 & \textbf{1530} & 30 & 300 \\
Name This Game      & \textbf{1400} & 720 & 190                  & 425 & \textbf{781} & 396 & 600 \\
Phoenix             & \textbf{570} & 255 & 35                    & 12 & \textbf{330} & 25.5 & 200 \\
Q*Bert              & \textbf{1150} & 220 & 105                  & 120 & \textbf{125} & 83 & 140 \\
Road Runner         & \textbf{10000} & 330 & 500                 & 236 & \textbf{354} & 187 & 180 \\
Atlantis            & \textbf{4680} & 980 & 760                  & \textbf{1650} & \textbf{1650} & 410 & 120 \\
Centipede           & \textbf{1500} & 1300 & 370                 & \textbf{900} & \textbf{900} & 227 & 250 \\
Bank Heist          & \textbf{185} & 2 & 2.5                     & \textbf{16.6} & 1 & 3.8 & 300 \\
Beam Rider          & \textbf{360} & 210 & 143                   & \textbf{165} & 140 & 65 & 600 \\
Boxing              & \textbf{10} & -26 & -26                    & \textbf{-8.7} & -26 & -35 & 1800 \\
Gopher              & \textbf{640} & 145 & 41                    & \textbf{298} & 42 & 42 & 350 \\
Gravitar            & \textbf{30} & 20 & 21.2                    & \textbf{50} & 19 & 2.2 & 125 \\
Krull               & \textbf{1725} & 725 & 330                  & \textbf{1745} & 398 & 140 & 500 \\
Ms Pac-Man          & \textbf{590} & 190 & 190                   & \textbf{348} & 235 & 175 & 210 \\
Pitfall!            & \textbf{-3.8} & -16 & -37.2                & \textbf{-9.8} & -15.4 & -12.3 & 700 \\
Pong                & \textbf{19.5} & -20.8 & -17                & \textbf{-18.5} & -20.9 & -20.7 & 1000 \\
River Raid          & \textbf{1160} & 490 & 255                  & \textbf{513} & 381 & 150 & 150 \\
Robotank            & \textbf{1.85} & 1.8 & 0.8                  & \textbf{2.43} & 0.8 & 1.2 & 600 \\
Video Pinball       & \textbf{11600} & 4440 & 1000               & \textbf{3000} & 2500 & 330 & 800 \\
Alien               & \textbf{475} & 121 & 140                   & \textbf{150} & 110 & \textbf{150} & 200 \\
HERO                & \textbf{3525} & 3520 & 1765                & 1160 & 1090 & \textbf{1960} & 400 \\
Assault             & \textbf{400} & 140 & 53                    & 83.4 & 73 & \textbf{97} & 200 \\
Demon Attack        & \textbf{580} & 370 & 110                   & 53 & 54 & \textbf{110} & 1000 \\
Private Eye         & \textbf{1200} & 600 & 700                  & 155 & 10 & \textbf{1500} & 2700 \\
Time Pilot          & 290 & \textbf{460} & 330                   & 98 & \textbf{393} & 221 & 300 \\
Up and Down         & 515 & \textbf{700} & 258                   & 350 & \textbf{592} & 160 & 150 \\
James Bond          & 32.3 & \textbf{39.5} & 16                  & 9.5 & \textbf{15.8} & 1.4 & 120 \\
Berzerk             & 150 & \textbf{900} & 125                   & 73 & \textbf{128} & 100 & 150 \\
Bowling             & 30 & \textbf{30.8} & 27                    & 20 & \textbf{30} & 21 & 2350 \\
Yars' Revenge       & 1337 & \textbf{2005} & 900                 & 1300 & \textbf{1950} & 900 & 160 \\
Solaris             & 292 & \textbf{370} & 116                   & \textbf{250} & 210 & 45 & 2500 \\
Frostbite           & 100 & \textbf{105} & 50                    & \textbf{50} & 48.7 & 25 & 150 \\

\midrule

Zaxxon              & 0 & 0 & 0                         & 0 & 0 & 0 & 175 \\
Venture             & 0 & 0 & 0                         & 0 & 0 & 0 & 700 \\
Ice Hockey          & -14 & -14 & -15                   & -11 & -12.7 & -15 & 3400 \\
Double Dunk         & -22.7 & -22.1 & -22.5             & -20.5 & -21.8 & -22.5 & 6000 \\
Tennis              & -23.3 & -23.8 & -23.9             & -22.8 & -23.8 & -23.9 & 6000 \\
Fishing Derby       & -89 & -91 & -90.8                 & -91 & -94.1 & -91 & 1850 \\
Skiing              & -31,000 & -31000 & -31,000        & -31000 & -31000 & -31000 & 4400 \\
Chopper Command     &  & &                              &  &  &  & 16000 \\
Tutankham           &  & &                              &  &  &  & 4500 \\
Enduro              &  & &                              &  &  &  & 3320 \\

\bottomrule
\end{tabular}
\end{sc}
\end{small}
\end{center}
\end{table*}

\section{Hyper-parameters}

Table \ref{hyperparameters} shows the hyper-parameters for the four experiments in this work. The dashes indicate that the hyper-parameter was not used, either because the training was performed offline or because of a design decision.

\begin{table*}[b]
\caption{Hyper-parameters used for the different experiments.}
\label{hyperparameters}

\begin{center}
\begin{small}
\begin{tabular}{|l|l|l|l|l|l|}
    \toprule
    Parameter Name          & GW Online        & GW Offline        & Healthcare    & CartPole      & Atari \\
    \midrule
    InferNet Hidden Layers  & 3 Dense          & 3 Dense           & 3 Dense       & 3 Dense       & 3 Conv + 1 Dense      \\
    InferNet Num. Units     & 3x256            & 3x256             & 3x256         & 3x64          & Conv: 32, 64, 64. Dense: 512      \\
    InferNet Activation     & Leaky ReLU       & Leaky ReLU        & Leaky ReLu    & ReLU          & ReLU      \\
    InferNet Dropout Rate   & ---              & ---               & 0.2           & 0.2           & 0.2      \\
    InferNet Optimizer      & Adam             & Adam              &    Adam       & Adam          & Adam      \\
    InferNet Learning Rate  & 1e-4             & 1e-3              &   1e-4        & 1e-4          & 3e-3      \\
    InferNet Batch Size     & 32 ep.           & 32 ep.            &   20 ep.      & 10 ep.        & 1 ep.      \\
    InferNet Training Steps & 500,000          & 50,000            &   1,000,000   & 60,000        & Varying per game     \\
    InferNet Buffer Size    & 500              & ---               & ---           & 500 ep.       & 500 ep.      \\
    Agent Training Steps    & 2,000 ep.        & 5,000             & 1,000,000     & 150,000       & Varying per game      \\
    Agent Discount $\gamma$ & 0.9              & 0.90              & 0.99          & 0.99          & 0.99      \\
    Agent Batch Size        & ---              & 32                & 32            & 32            & 32      \\
    Agent Buffer Size       & ---              & ---               & ---           & 500,000       & 10,000      \\
    Agent Hidden Layers     & ---              & ---               & 2 Dense       & 2 Dense       & 3 Conv + 1 Dense      \\
    Agent Num. Units        & ---              & ---               & 2x256         & 2x32          & Conv: 32, 64, 64. Dense: 512      \\
    Agent Activation        & ---              & ---               & ReLU          & ReLU          & ReLU      \\
    Agent Learning Rate     & ---              & ---               & 1e-4          & 2.5e-4        & 1e-4      \\
    TD($\lambda$): $\lambda$ & 0.91            & ---               & ---           & ---           & ---      \\
    TD($\lambda$): $\alpha$ & 0.1              & ---               & ---           & ---           & ---      \\
    TD($\lambda$): traces   & Dutch            & ---               & ---           & ---           & ---      \\
    \bottomrule
\end{tabular}
\end{small}
\end{center}

\end{table*}

\section{Ethical Considerations}
In our human-machine mixed initiative RL framework, human experts are the final decision makers, and the RL agent assists for them to make a better and timely decision. In healthcare,
despite the vast amount of data used to induce RL policies, the types of input are limited to EHRs and do not cover all the considerations as human physicians do in terms of knowledge, resources, finance, patients' preference, and unrecorded context. With such limitations, the RL agent could still help domain experts as an assistant to timely suggest best possible treatment options with their expected consequences by learning and generalizing an effective treatment policy from the medical records, related to similar disease or symptoms. Such an application would particularly benefit medical students or busy experts in emergency rooms requiring urgent decisions, and many hospitals suffering from a lack of expertise on a specific disease such as sepsis. 

\end{document}